\documentclass[runningheads]{llncs}
\usepackage[T1]{fontenc}
\usepackage{amsfonts}
\usepackage{amsmath}
\usepackage{hyperref}
\usepackage{booktabs}
\usepackage{color}
\usepackage{hyperref}
\usepackage[table,xcdraw]{xcolor}
\usepackage{bbding}
\usepackage{graphicx,verbatim}
\begin{document}
    \title{A Causality-Inspired Model for Intima-Media Thickening Assessment in Ultrasound Videos}
%
\begin{comment}  %% Removed for anonymized MICCAI 2025 submission
\author{First Author\inst{1}\orcidID{0000-1111-2222-3333} \and
Second Author\inst{2,3}\orcidID{1111-2222-3333-4444} \and
Third Author\inst{3}\orcidID{2222--3333-4444-5555}}
%
\authorrunning{F. Author et al.}
\institute{Princeton University, Princeton NJ 08544, USA \and
Springer Heidelberg, Tiergartenstr. 17, 69121 Heidelberg, Germany
\email{lncs@springer.com}\\
\url{http://www.springer.com/gp/computer-science/lncs} \and
ABC Institute, Rupert-Karls-University Heidelberg, Heidelberg, Germany\\
\email{\{abc,lncs\}@uni-heidelberg.de}}

\end{comment}

\vspace{-10pt}
\author{
    Shuo Gao\textsuperscript{\textdagger1}, 
    Jingyang Zhang\textsuperscript{\textdagger2(\Envelope)}, 
    Jun Xue\textsuperscript{3}, 
    Meng Yang\textsuperscript{3}, 
    Yang Chen\textsuperscript{2}, and Guangquan Zhou\textsuperscript{1(\Envelope)}
}

\authorrunning{Shuo Gao et al.}  % 这里会在页眉显示简化的作者名

\institute{
    \textsuperscript{1}School of Biological Science and Medical Engineering, Southeast University, \\
    Nanjing, China \\
    \email{guangquan.zhou@seu.edu.cn} \\
    \textsuperscript{2}School of Computer Science and Engineering, Southeast University, \\
    Nanjing, China \\
    \email{j.y.zhang@seu.edu.cn} \\
    \textsuperscript{3}Peking Union Medical College Hospital, Chinese Academy of Medical Sciences, \\
    Beijing, China
}

\maketitle              % typeset the header of the contribution
\vspace{-10pt}
\begin{abstract}
Carotid atherosclerosis represents a significant health risk, with its early diagnosis primarily dependent on ultrasound-based assessments of carotid intima-media thickening. However, during carotid ultrasound screening, significant view variations cause style shifts, impairing content cues related to thickening, such as lumen anatomy, which introduces spurious correlations that hinder assessment. Therefore, we propose a novel causal-inspired method for assessing carotid intima-media thickening in frame-wise ultrasound videos, which focuses on two aspects: eliminating spurious correlations caused by style and enhancing causal content correlations. Specifically, we introduce a novel Spurious Correlation Elimination (SCE) module to remove non-causal style effects by enforcing prediction invariance with style perturbations. Simultaneously, we propose a Causal Equivalence Consolidation (CEC) module to strengthen causal content correlation through adversarial optimization during content randomization. Simultaneously, we design a Causal Transition Augmentation (CTA) module to ensure smooth causal flow by integrating an auxiliary pathway with text prompts and connecting it through contrastive learning. The experimental results on our in-house carotid ultrasound video dataset achieved an accuracy of 86.93\%, demonstrating the superior performance of the proposed method. Code is available at \href{https://github.com/xielaobanyy/causal-imt}{https://github.com/xielaobanyy/causal-imt}.

% Authors must provide keywords and are not allowed to remove this Keyword section.

\end{abstract}

\let\thefootnote\relax
\footnotetext{\textsuperscript{\textdagger} Equal contribution; \textsuperscript{\Envelope} Corresponding author.}
\section{Introduction}

\begin{figure}
\includegraphics[width=\textwidth]{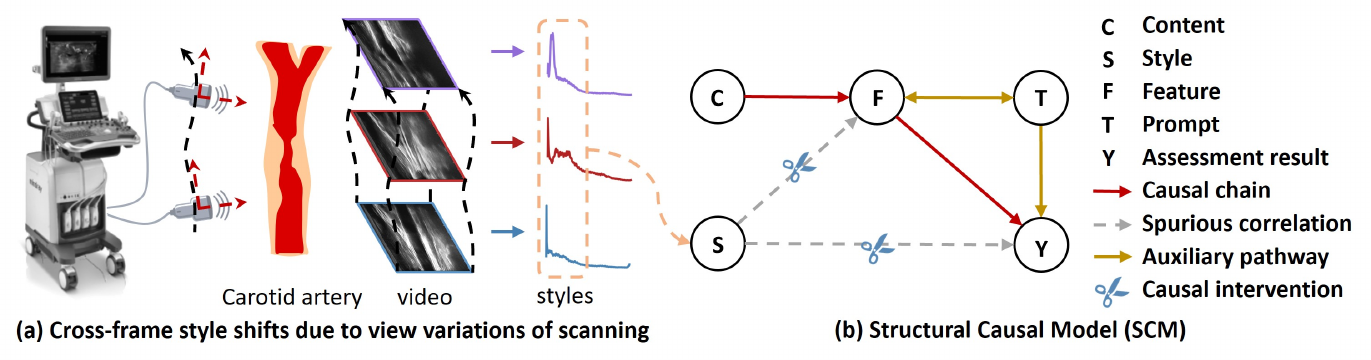}
\setlength{\abovecaptionskip}{-0.3cm}
\caption{The causal mechanism for intima-media assessment: (a) Cross-frame style shifts due to view variations of scanning; (b) Structural Causal Model (SCM).} \label{fig1}
\end{figure}

Carotid atherosclerosis is a widespread and complex cardiovascular disease that poses a notable global public health threat \cite{bir2022carotid}.
Intima-Media Thickening (IMT) acts as a critical indicator for screening carotid atherosclerosis \cite{nezu2016carotid}, where ultrasound emerges as the preferred modality for imaging owing to its non-invasiveness and accessibility \cite{polak2016carotid}.
In clinical practice, the physician maneuvers the ultrasound prob along the carotid artery to collect a complete ultrasound video for tracking intima-media changes.
This video contains numerous frames that reveal the spatial distribution of thickening along the carotid artery \cite{espeland1994spatial}, providing valuable insights for characterizing vascular health and designing personalized treatment.
Therefore, it is desirable to perform IMT assessment in a frame-wise manner for the carotid ultrasound video, leading to fine-grained cues for thickening position.

For frame-wise IMT assessment in carotid ultrasound videos, existing methods \cite{he2024deep,singh2024atherosclerotic} typically treat it as a frame-level event detection task \cite{chen2022frame}, often relying on off-the-shelf paradigms from the field of generic video processing.
For example, the spatiotemporal aggregation module originally designed for generic object tracking \cite{cheng2022xmem} is directly transferred to the carotid ultrasound video for capturing dynamic IMT features \cite{shan2024lsmd}.
As these existing methods are rooted for the general video analysis with only mild view variations across frames \cite{fasching2015classification},
their efficacy on carotid ultrasound videos for IMT assessment heavily relies on a strict scanning protocol \cite{touboul2012assessment},
i.e., holding the ultrasound prob with a fixed position and static orientation to keep cross-frame view stability.
However, this protocol contradicts clinical practice, where the physician skillfully maneuvers the ultrasound probe, dynamically scanning along the carotid artery while adjusting its orientation to optimize tissue penetration with clear intima-media \cite{landwehr2001ultrasound}, as shown in Fig. \ref{fig1}(a).
In this way, substantial view variations would occur across frames with discrepancies in echo intensity, causing style shifts that corrupt IMT-related content cues, e.g., lumen anatomy.
Hence, IMT assessment for these view-varying frames tends to be misguided by spurious correlations to style shifts, while ignoring causal correlations to content cues.

To address this issue, we construct a Structural Causal Model (SCM) for IMT assessment, as shown in Fig. \ref{fig1}(b), to identify the non-causal impact
pathway of spurious correlations and explore an ideal causal chain.
Theoretically, the causal content $C$ should be the only endogenous parent that determines IMT assessment $Y$ by deriving the feature $F$ as an intermediate result, thereby forming a causal chain $C \rightarrow F \rightarrow Y$.
However, view-varying frames contains diverse styles $S$ that misguide feature extraction and, finally, IMT assessment through $S \rightarrow F \rightarrow Y$,
opening a backdoor path with spurious correlation $S \dashrightarrow Y$.
Based on the above causality mechanism, the elimination of spurious correlation can be achieved by blocking the non-causal path, i.e., $S \not\rightarrow F \rightarrow Y$, shielding prediction from influence of style shifts.
Moreover, causal correlations can be enhanced via the causal chain $C \rightarrow F \rightarrow Y$ with two aspects:
1) ensuring causal equivalence between $C$ and $Y$ for a direct impact consolidation;
2) improving causal transition by augmenting an auxiliary pathway $C \rightarrow F \leftrightarrow T \rightarrow Y$, where $T$ prompts $Y$ and recalibrate $F$ to smoothly transfer causal impact through the causal chain.

In this paper, we present, to our knowledge, \emph{the first causality-inspired model} for IMT assessment in ultrasound videos.
This model eliminates spurious style correlations across view-varying frames while enhancing the causal content correlations for accurate assessment.
Specifically, our contributions are three folds:
1) We propose a Spurious Correlation Elimination (SCE) module that cut-offs non-causal style impacts by enforcing prediction invariance with simulated style perturbations;
2) We develop a Causal Equivalence Consolidation (CEC) module to enhance the direct causal content correlation, via adversarial optimization on assessment predictions under content randomization;
and 3) We design a Causal Transition Augmentation (CTA) module for a smooth causal impact flow, where an auxiliary causal pathway is formed using text prompts with chain-of-thought guidance and further involved into the causal chain by contrastive learning.
We evaluate our method on a in-house dataset of carotid ultrasound videos, showing its highest accuracy and clear advantages for frame-wise IMT assessment.

\section{Methodology}
\begin{figure}
\includegraphics[width=\textwidth]{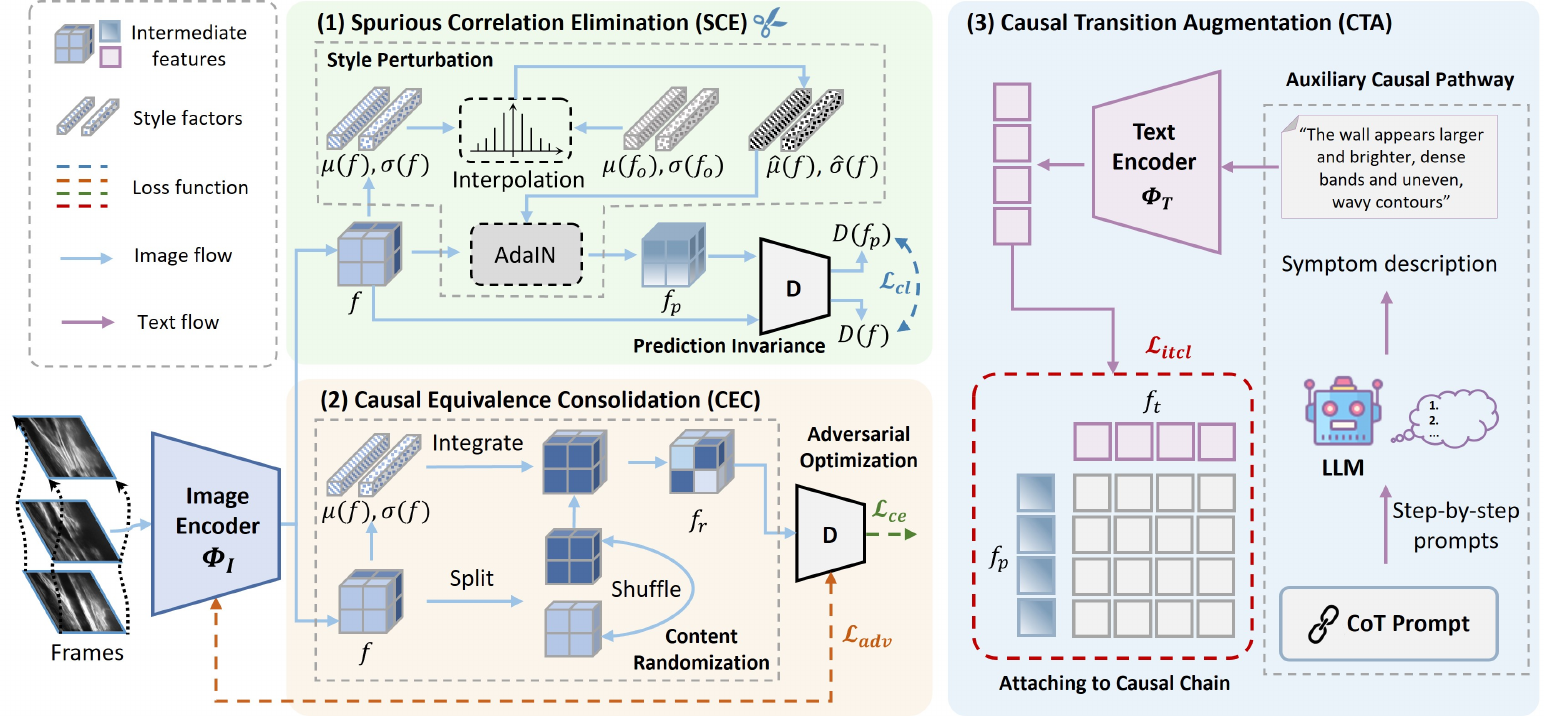}
\caption{The architecture of the proposed method. Specifically, we apply Spurious Correlation Elimination (SCE) module to remove non-causal style effects by enforcing prediction invariance with style perturbations (Sect. \ref{SCE}), then we design Causal Equivalence Consolidation (CEC) module to strengthen causal content correlation by adversarial optimization during content randomization (Sect. \ref{CEC}). Furthermore, we design a Causal Transition Augmentation (CTA) module to enhance causal effects by integrating an auxiliary causal pathway and linking it by contrastive learning (Sect. \ref{CTA}).} \label{fig2}
\end{figure}

As illustrated in Fig.~\ref{fig2}, the frames of a carotid ultrasound video are fed into the proposed causality-inspired model for frame-wise IMT assessment.
Specifically, the style shifts across these view-varying frames are identified as spurious correlations to be eliminated, while causal content factors are enhanced via equivalence consolidation and transition augmentation.

\subsection{Spurious Correlation Elimination (SCE)}
\label{SCE}

In clinical practice, physicians flexibly manipulate the ultrasound prob for carotid scanning, leading to varying views across frames in a ultrasound video. It would cause cross-frame shifted styles $S$, which corrupts the extracted features $F$ and finally misguides the IMT assessment $C$ via a spurious correlation ($S \rightarrow F \rightarrow Y$). To eliminate such spurious correlation, we apply a causal intervention on $F$ by a do-operator \cite{jiao2024causal}, which simplifies the correlation analysis between $S$ and $Y$ with $F$ conditional fixed:
\begin{equation}
\label{formula-1}
P(Y|do(F)) = {\sum}_{S} P(Y|do(F), S) \cdot P(S|do(F)) = {\sum}_{S} P(Y|F, S) \cdot P(S),
\end{equation}

\noindent
which depicts the cumulative effect of multiple $S$ on $Y$.
It implies that, to eliminate their spurious correlations, it is crucial to disentangle the correlations between $Y$ and a sufficiently large set of $S$.
To this end, we simulate style perturbations with a highly diverse distribution and enforce prediction invariance over them, facilitating potential spurious correlations can be thoroughly eliminated.

\vspace{0.3em}
\noindent
\textbf{Style Perturbation.}
Style factor is typically represented using feature statistic, i.e., mean $\mu$ and standard deviation $\sigma$ \cite{liu2024causality}. To ensure a wide perturbation range for style, we leverage shifted styles $\mu(f_o)$ and $\sigma(f_o)$ from view-varying frames and further augment them via random style interpolation, leading to a perturbed style $\hat{\mu}(f)$ and $\hat{\sigma}(f)$:
%To achieve style augmentation, we interpolate style representations across different frames and then perform normalization on features with augmented styles. Specifically, the style factor is represented using intermediate feature statistic \cite{liu2024causality} from the video frames (mean and standard deviation, $\mu$, $\sigma$ $\in$ $\mathbb{R}^C$) and further augmented through random interpolation.
\begin{equation}
\hat{\mu}(f) = \lambda \cdot \mu(f) + (1 - \lambda) \cdot \mu(f_o), \, \hat{\sigma}(f) = \lambda \cdot \sigma(f) + (1 - \lambda) \cdot \sigma(f_o),
\end{equation}
where $\lambda$ $\sim$ Uniform(0, 1) is a random interpolation weight. After acquiring this diverse style, we further transfer it by Adaptive Instance Normalization (AdaIN) \cite{huang2017arbitrary} to obtain style-perturbed features $f_p$ based on original features $f$:

\begin{equation}
f_p = {\hat{\sigma}(f)} \cdot \left( \frac{f - \mu(f)}{\sigma(f)} \right) + {\hat{\mu}(f)}.
\end{equation}

\vspace{0.3em}
\noindent
\textbf{Prediction Invariance.}
% \subsubsection{Consistency Constraint on Assessment.}
Eventually, we design a consistency loss $\mathcal{L}_{\text{cl}}$ using KL divergence $D_{KL}$ to ensure the invariance of predictions between $f$ and $f_p$:

\begin{equation}
\min_{\Phi_I,\,D} \mathcal{L}_{\text{cl}}= D_{{KL}}[D(f) || D(f_p)] + D_{{KL}}[D(f_p) || D(f)],
\end{equation}
\begin{equation}
\min_{D}\mathcal{L}_{\text{ce}} = - y\log \left( D\left( f \right) \right).
\end{equation}

\noindent
where \( D(f) \) and \( D(f_p) \) represent the corresponding predicted distributions, and $y$ represents the labels for thickening or non-thickening.

\subsection{Causal Equivalence Consolidation (CEC)}
\label{CEC}

In the idea causal chain, the assessment prediction $Y$ is primarily determined by the content $C$ via the intermediate feature $F$, i.e., $C \rightarrow F \rightarrow Y$.
It motivates that the prediction results should be strongly correlated with the content factor, i.e., with high causal equivalence.
To achieve this, we propose a Causal Equivalence Consolidation (CEC) module to strengthen the causal content correlation, which randomizes content factors and performs adversarial optimization on assessment predictions to enforce their causal equivalence towards content.

%Additionally, according to the SCM,  to further reinforce the causal relationship ($C \rightarrow F_{I} \rightarrow Y$), we hypothesize that the primary content information is encoded across different channels of the image features. By randomizing the channel order of the feature, we can effectively simulate significant changes in the content information. Specifically, the intermediate feature $f_I$ is split and randomly shuffled along the channel dimension, then combined with the style features $\mu(f_I)$ and $\sigma(f_I)$ to generate perturbed image features $f_I^{\prime}$. Furthermore, we aim to mitigate the interference of style factors by constructing an adversarial structure. The perturbed image features are evaluated by the style discriminator $D$ to classify them, while ensuring that the image features produced by $\Phi{_I}$ can effectively deceive $D$ after intervention.

\vspace{0.3em}
\noindent
\textbf{Content Randomization.}
As the content factor is typically defined as the channel order of the feature map \cite{gatys2017controlling}, we randomize the content by independently splitting each feature channel and shuffling them, achieving features with random content $\mathcal{R}(f)$.
Moreover, to avoid potential style degeneration during this content randomization \cite{deng2022stytr2}, we integrate the original style factors $\mu{(f)}$ and $\sigma{(f)}$ into $\mathcal{R}(f)$ for style stability.
\begin{equation}
f_r = {\mathcal{R}(f)} \cdot \sqrt{{\sigma(f)}^2 + \epsilon} + \mu(f).
\end{equation}
\noindent

\vspace{0.3em}
\noindent
\textbf{Adversarial Optimization.}
To ensure causal equivalence between content and assessment predictions, we employ adversarial optimization on these assessment predictions to align their distribution with even randomized contents.
Specifically, we minimize the adversarial loss $\mathcal{L}_{\text{adv}}$ to encourage the image encoder $\Phi_I$ to fool the classifier $D$, making it use content-randomized feature $f_r$ to generate low-confidence predictions $D(f_r)$ with uniform probability $U=1/2$:

\begin{equation}
\min_{\Phi_I}\mathcal{L}_{\text{adv}} = - U \log \left( D\left( f_r \right) \right).
\end{equation}

\noindent
Intuitively, randomized contents are enforced to produce meaningless assessment, enhancing the causal sensitivity to the content and ensuring causal equivalence.

\subsection{Causal Transition Augmentation (CTA)}
\label{CTA}

Besides causal quaivalence consolidation, another aspect for causal correlation enhancement is to design a Causal Transition Augmentation (CTA) mechanism, where an auxiliary pathway is attached upon the causal chain to avoid the degraded causal propagation, i.e., $C \rightarrow F \leftrightarrow T \rightarrow Y$.
For this goal, we construct an auxiliary causal pathway using text prompts guided by chain-of-thought for fine-grained IMT symptoms. This pathway is then attached to the original causal chain using contrastive learning for alignment.

\begin{figure}[t]
\includegraphics[width=\textwidth]{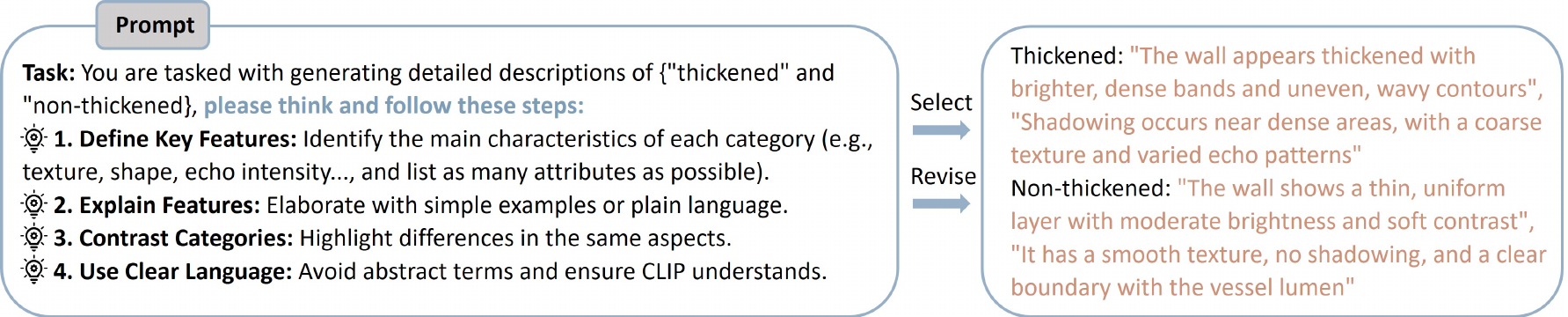}
\caption{Generation of intima-media symptom descriptions using step-type prompts.} \label{fig3}
\end{figure}

\vspace{0.3em}
\noindent
\textbf{Auxiliary Causal Pathway}
Text conveys abundant information that complements images \cite{long2022gradual}, serving as a valuable data resource for constructing an auxiliary causal pathway.
However, naive text prompts are not well-suited for IMT assessment, as it is a highly complex task that typically demands a step-by-step logical process to define key symptoms.
Motivated by this, we leverage the chain of thought \cite{wei2022chain} mechanism to guide the generation of fine-grained text prompts, as shown in Fig.~\ref{fig3}.
Such text prompts with symptom description establish an auxiliary pathway that augments the original image-based causal chain.

\vspace{0.3em}
\noindent
\textbf{Attaching to Causal Chain.}
Since the auxiliary causal pathway is constructed using text prompts, we adpot contrastive learning \cite{zhang2022contrastive} to align these text prompts with images into a shared feature space, attaching it to the original image-based causal chain.
Specifically, text prompts are fed into the text encoder $\Phi_T$ to derive the text feature $f_t$, which is aligned to the image feature $f$ via a contrastive loss:

\begin{equation}
\min_{\Phi_I} \mathcal{L}_{\text{itcl}} = - \log \frac{\exp\left(f \cdot f_t^+/ \tau\right )} {\exp\left(f \cdot f_t^+ / \tau\right ) + \sum_{i \in \mathcal{N}} \exp\left(f \cdot f_t^{(i)} / \tau\right)}.
\end{equation}
\noindent
where $f_t^+$ is the embedding of the positive symptom prompt, $\mathcal{N}$ represents the set of negative samples and $f_t^{(i)}$ represents the negative ones. $(\cdot)$ denotes the cosine similarity between two embeddings. $\tau$ is a temperature scaling parameter.

\subsection{Overall Loss}
The total loss of our proposed causality-inspired model consists of several parts, where $\alpha_{1}$ , $\alpha_{2}$, and $\alpha_{3}$ are trade-off parameters.
\begin{equation}
\min_{\Phi_I}\mathcal{L}_{\text{itcl}}+
\min_{\Phi_I,\,D}\alpha_{1} \mathcal{L}_{\text{cl}}+ \min_{D}\alpha_{2}\mathcal{L}_{\text{ce}}+ \min_{\Phi_I}\alpha_{3}\mathcal{L}_{\text{adv}}.
\end{equation}

\section{Experiments}

\vspace{0.3em}
\noindent
\textbf{Datasets.} In the IMT assessment task, we used a carotid ultrasound video dataset comprising 120 videos. All frames were cropped to 506×477, resized to 224×224, and categorized into two types: thickening and non-thickening. Each frame is paired with a fine-grained textual description of its category and additional frame data for style information. The dataset was split into 60\%, 20\%, and 20\% for training, validation, and testing, respectively. For this dataset, we adopt Accuracy, Sensitivity, Sensitivity, Precision and F1-score as evaluation metrics.

\begin{table}
\centering
\caption{The quantitative evaluation demonstrates the superiority of our method.}\label{tab1}
\begin{tabular}{ccccc}
\toprule
Method & Accuracy(\%) & Sensitivity(\%) & Precision(\%)  & F1-Score(\%) \\
\midrule
ResNet-50 \cite{he2016deep}    & 78.14        & 32.35           & 76.64         & 77.30 \\
APCNet \cite{singh2024atherosclerotic}      & 77.60        & 23.53           & 74.65         & 75.83 \\
DCCNet \cite{yang2024automated}      & 70.49        & 8.82            & 67.49         & 68.91 \\
LSMD \cite{shan2024lsmd}             & 81.42        & 21.46           & 40.71         & 44.88 \\
TSM \cite{lin2019tsm}              & 77.77        & 14.84           & 85.19         & 77.78 \\
MVFNet \cite{wu2021mvfnet}           & 75.00        & 85.71           & 89.28         & 85.89 \\
Proposed          & \pmb{86.93}  & \pmb{88.89}     & \pmb{91.10}   & \pmb{88.10} \\
\bottomrule
\end{tabular}
\end{table}

\begin{table}
\centering
\caption{The ablation results of the proposed module.}\label{tab2}
\begin{tabular}{ccccccc}
\toprule
SCE       &      CEC &     CTA   & Accuracy(\%) & Sensitivity(\%)&Precision(\%)    & F1-Score(\%) \\
\midrule
\checkmark & \checkmark&             & 84.65        & 21.33           & 71.67                & 77.62 \\
           & \checkmark&  \checkmark & 76.70        & 51.85          & 81.56                & 78.62 \\
\checkmark &           &  \checkmark & 85.23        & 88.89          & 90.52                & 86.69 \\
\checkmark & \checkmark& \checkmark  & \pmb{86.93}  & \pmb{88.89}    & \pmb{91.10}    & \pmb{88.10} \\
\bottomrule

\end{tabular}
\end{table}

\vspace{0.3em}
\noindent
\textbf{Implementation Details.} We employed the pre-trained CLIP (ViT-B/16) model for image and text encoding \cite{radford2021learning}. It was trained using SGD Optimizer with learning rate $6 \times 10^{-5}$, batch size 16, and epoch number 100. Beside, We empirically set the overall loss weight coefficients $\alpha_1$, $\alpha_2$, and $\alpha_3$ to 0.5, 0.1, and 0.1, respectively.

\begin{figure}[t]
\centering
\includegraphics[width=0.8\textwidth]{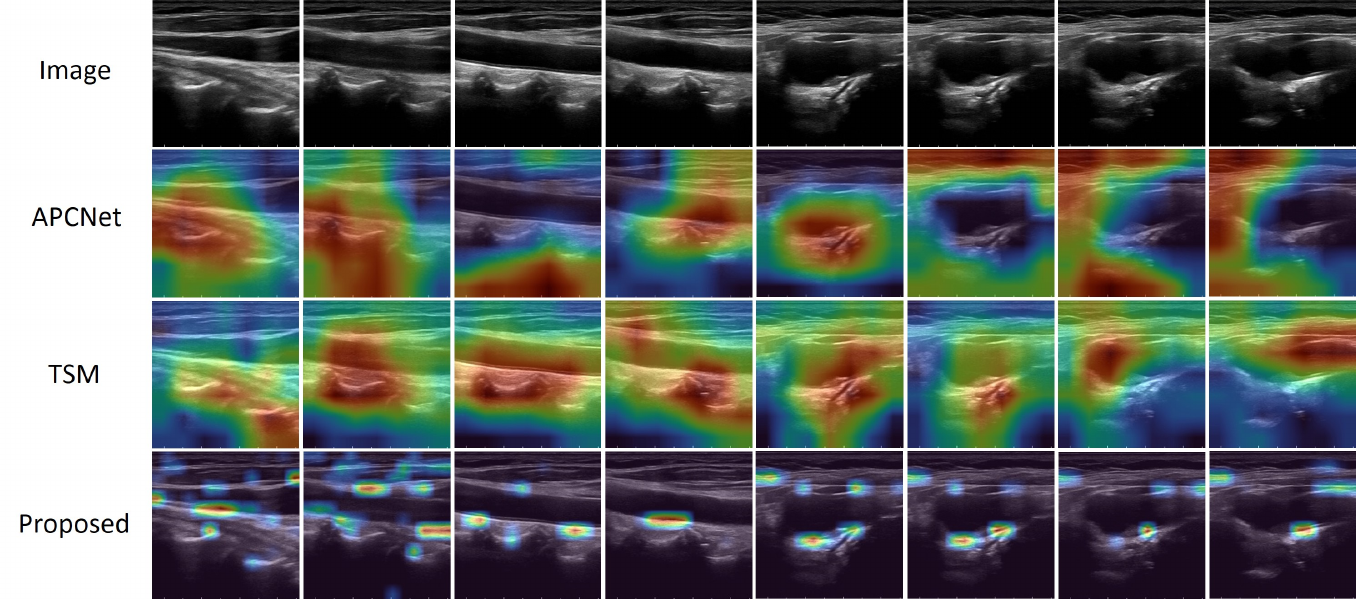}
\caption{Grad-CAM visualization results for different methods. The first row shows the original video frames, with the first four representing non-thickened frames and the last four representing thickened frames.} \label{fig4}
\end{figure}

\vspace{0.3em}
\noindent
\textbf{Comparison with State-of-the-Arts.} We compare our method with existing approaches, such as \textbf{ResNet} \cite{he2016deep}, \textbf{APCNet}\cite{singh2024atherosclerotic}, \textbf{DCCNet}\cite{yang2024automated}, \textbf{LSMD} \cite{shan2024lsmd}, \textbf{TSM} \cite{lin2019tsm} and \textbf{MVFNet} \cite{wu2021mvfnet}. Competing models are retrained on our datasets using the training codes provided by their respective authors. The results indicate that our model shows superior performance in four key aspects (Table~\ref{tab1}). Proposed method achieve superior performance in all experiments, surpassing the next best method, MVFNet, with average improvements of 11.93$\%$, 3.18$\%$, 1.81$\%$, and 2.21$\%$ in accuracy, sensitivity, precision, and f1-score, respectively. Notably, due to the limited number of positive thickening cases, most models pose significant long-tail challenges during training. For example, the sensitivity for positive thickening in ResNet, APCNet, and DCCNet is 32.35, 23.53, and 8.82, respectively. However, our proposed method effectively addresses this issue. This success can be attributed to its ability to eliminate the spurious correlations caused by style shifts and enhance causal correlations to content, allowing the model to rely on content information to make decisions across different frames. Fig.~\ref{fig4} shows the visualization results of Grad-CAM \cite{selvaraju2017grad}, which shows that our proposed model focuses on the content cues of the intima-media, such as bifurcation locations and intima-media structure (last row). Furthermore, by incorporating customized intima-media symptom texts, this method captures fine-grained information related to thickening, enabling it to distinguish subtle differences between the two categories.

\begin{figure}[t]
\centering
\includegraphics[width=0.8\textwidth]{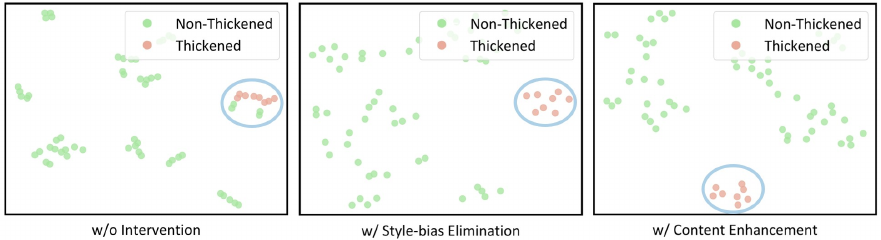}
\caption{t-SNE visualization in a batch with and without intervention.} \label{fig5}
\end{figure}

\vspace{0.3em}
\noindent
\textbf{Ablation Study.} Table~\ref{tab2} summarizes the ablation study for each module on the intima-media dataset. The results show that both the SCE and CEC significantly improve prediction performance. Specifically, incorporating SCE leads to improvements across all performance metrics, with accuracy and precision increasing by 10.23\% and 9.54\%, respectively. Furthermore, adding CEC further boosts model accuracy by 1.7\%. Fig.~\ref{fig5} shows the t-SNE results of image feature distributions with and without causal intervention. With the intervention, the two categories are clearly separated. Furthermore, compared to descriptions generated by general prompt-based methods (\emph{Large, irregular zones dominate the image, with uneven borders and rough lines}), CTA demonstrates more robust performance in handling imbalanced class cases.

\section{Conclusion}
The assessment of IMT plays a vital role in the diagnosis and treatment of carotid atherosclerosis. To achieve this, we propose an assessment method based on carotid ultrasound videos from a causal perspective. The method eliminates interference from style variations between video frames through causal modeling. Additionally, it enhances robustness by incorporating fine-grained symptom descriptions. This approach offers a new perspective for IMT assessment and contributes to advancing the application of causal learning in cardiovascular disease research. In the future, we will conduct more comprehensive experiments to validate the robustness of this method.

\bibliographystyle{splncs04}
\bibliography{ref/imt}

\begin{thebibliography}{10}
\providecommand{\url}[1]{\texttt{#1}}
\providecommand{\urlprefix}{URL }
\providecommand{\doi}[1]{https://doi.org/#1}

\bibitem{bir2022carotid}
Bir, S.C., Kelley, R.E.: Carotid atherosclerotic disease: A systematic review of pathogenesis and management. Brain circulation  \textbf{8}(3),  127--136 (2022)

\bibitem{chen2022frame}
Chen, M., Wei, F., Li, C., Cai, D.: Frame-wise action representations for long videos via sequence contrastive learning. In: Proceedings of the IEEE/CVF Conference on Computer Vision and Pattern Recognition. pp. 13801--13810 (2022)

\bibitem{cheng2022xmem}
Cheng, H.K., Schwing, A.G.: Xmem: Long-term video object segmentation with an atkinson-shiffrin memory model. In: European Conference on Computer Vision. pp. 640--658. Springer (2022)

\bibitem{deng2022stytr2}
Deng, Y., Tang, F., Dong, W., Ma, C., Pan, X., Wang, L., Xu, C.: Stytr2: Image style transfer with transformers. In: Proceedings of the IEEE/CVF conference on computer vision and pattern recognition. pp. 11326--11336 (2022)

\bibitem{espeland1994spatial}
Espeland, M.A., Hoen, H., Byington, R., Howard, G., Riley, W.A., Furberg, C.D.: Spatial distribution of carotid intimal-medial thickness as measured by b-mode ultrasonography. Stroke  \textbf{25}(9),  1812--1819 (1994)

\bibitem{fasching2015classification}
Fasching, J., Walczak, N., Morellas, V., Papanikolopoulos, N.: Classification of motor stereotypies in video. In: 2015 IEEE/RSJ International Conference on Intelligent Robots and Systems (IROS). pp. 4894--4900. IEEE (2015)

\bibitem{gatys2017controlling}
Gatys, L.A., Ecker, A.S., Bethge, M., Hertzmann, A., Shechtman, E.: Controlling perceptual factors in neural style transfer. In: Proceedings of the IEEE conference on computer vision and pattern recognition. pp. 3985--3993 (2017)

\bibitem{he2016deep}
He, K., Zhang, X., Ren, S., Sun, J.: Deep residual learning for image recognition. In: Proceedings of the IEEE conference on computer vision and pattern recognition. pp. 770--778 (2016)

\bibitem{he2024deep}
He, L., Yang, Z., Wang, Y., Chen, W., Diao, L., Wang, Y., Yuan, W., Li, X., Zhang, Y., He, Y., et~al.: A deep learning algorithm to identify carotid plaques and assess their stability. Frontiers in Artificial Intelligence  \textbf{7},  1321884 (2024)

\bibitem{huang2017arbitrary}
Huang, X., Belongie, S.: Arbitrary style transfer in real-time with adaptive instance normalization. In: Proceedings of the IEEE international conference on computer vision. pp. 1501--1510 (2017)

\bibitem{jiao2024causal}
Jiao, L., Wang, Y., Liu, X., Li, L., Liu, F., Ma, W., Guo, Y., Chen, P., Yang, S., Hou, B.: Causal inference meets deep learning: A comprehensive survey. Research  \textbf{7}, ~0467 (2024)

\bibitem{landwehr2001ultrasound}
Landwehr, P., Schulte, O., Voshage, G.: Ultrasound examination of carotid and vertebral arteries. European radiology  \textbf{11},  1521--1534 (2001)

\bibitem{lin2019tsm}
Lin, J., Gan, C., Han, S.: Tsm: Temporal shift module for efficient video understanding. In: Proceedings of the IEEE/CVF international conference on computer vision. pp. 7083--7093 (2019)

\bibitem{liu2024causality}
Liu, Y., Qin, G., Chen, H., Cheng, Z., Yang, X.: Causality-inspired invariant representation learning for text-based person retrieval. In: Proceedings of the AAAI Conference on Artificial Intelligence. vol.~38, pp. 14052--14060 (2024)

\bibitem{long2022gradual}
Long, S., Han, S.C., Wan, X., Poon, J.: Gradual: Graph-based dual-modal representation for image-text matching. In: Proceedings of the IEEE/CVF winter conference on applications of computer vision. pp. 3459--3468 (2022)

\bibitem{nezu2016carotid}
Nezu, T., Hosomi, N., Aoki, S., Matsumoto, M.: Carotid intima-media thickness for atherosclerosis. Journal of atherosclerosis and thrombosis  \textbf{23}(1),  18--31 (2016)

\bibitem{polak2016carotid}
Polak, J.F., O'Leary, D.H.: Carotid intima-media thickness as surrogate for and predictor of cvd. Global heart  \textbf{11}(3),  295--312 (2016)

\bibitem{radford2021learning}
Radford, A., Kim, J.W., Hallacy, C., Ramesh, A., Goh, G., Agarwal, S., Sastry, G., Askell, A., Mishkin, P., Clark, J., et~al.: Learning transferable visual models from natural language supervision. In: International conference on machine learning. pp. 8748--8763. PMLR (2021)

\bibitem{selvaraju2017grad}
Selvaraju, R.R., Cogswell, M., Das, A., Vedantam, R., Parikh, D., Batra, D.: Grad-cam: Visual explanations from deep networks via gradient-based localization. In: Proceedings of the IEEE international conference on computer vision. pp. 618--626 (2017)

\bibitem{shan2024lsmd}
Shan, C., Zhang, Y., Liu, C., Jin, Z., Cheng, H., Chen, Y., Yao, J., Luo, S.: Lsmd: Long-short memory-based detection network for carotid artery detection in b-mode ultrasound video streams. IEEE Transactions on Ultrasonics, Ferroelectrics, and Frequency Control  (2024)

\bibitem{singh2024atherosclerotic}
Singh, S., Jain, P.K., Sharma, N., Pohit, M., Roy, S.: Atherosclerotic plaque classification in carotid ultrasound images using machine learning and explainable deep learning. Intelligent Medicine  \textbf{4}(2),  83--95 (2024)

\bibitem{touboul2012assessment}
Touboul, P.J., Grobbee, D.E., Ruijter, H.d.: Assessment of subclinical atherosclerosis by carotid intima media thickness: technical issues. European journal of preventive cardiology  \textbf{19}(2\_suppl),  18--24 (2012)

\bibitem{wei2022chain}
Wei, J., Wang, X., Schuurmans, D., Bosma, M., Xia, F., Chi, E., Le, Q.V., Zhou, D., et~al.: Chain-of-thought prompting elicits reasoning in large language models. Advances in neural information processing systems  \textbf{35},  24824--24837 (2022)

\bibitem{wu2021mvfnet}
Wu, W., He, D., Lin, T., Li, F., Gan, C., Ding, E.: Mvfnet: Multi-view fusion network for efficient video recognition. In: Proceedings of the AAAI conference on artificial intelligence. vol.~35, pp. 2943--2951 (2021)

\bibitem{yang2024automated}
Yang, J., Li, X., Guo, Y., Song, P., Lv, T., Zhang, Y., Cui, Y.: Automated classification of coronary plaque on intravascular ultrasound by deep classifier cascades. IEEE Transactions on Ultrasonics, Ferroelectrics, and Frequency Control  (2024)

\bibitem{zhang2022contrastive}
Zhang, Y., Jiang, H., Miura, Y., Manning, C.D., Langlotz, C.P.: Contrastive learning of medical visual representations from paired images and text. In: Machine learning for healthcare conference. pp. 2--25. PMLR (2022)

\end{thebibliography}

\end{document}